\PassOptionsToPackage{table}{xcolor}
\documentclass[pdflatex,sn-mathphys-num]{sn-jnl}

\usepackage[dvipsnames]{xcolor}
\usepackage{graphicx}
\usepackage{lscape}
\usepackage[ruled,vlined]{algorithm2e}
\usepackage{amssymb}

\usepackage[table]{xcolor}
\usepackage{amsmath}
\usepackage{booktabs}

\usepackage{caption}%
\usepackage{subcaption}%

\theoremstyle{thmstyleone}%
\theoremstyle{thmstyletwo}%

\theoremstyle{thmstylethree}%

\begin{document}

\title[On the Fairness, Diversity and Reliability of Text-to-Image Generative Models]{On the Fairness, Diversity and Reliability of Text-to-Image Generative Models}


\author*[1]{\fnm{Jordan} \sur{Vice}}\email{jordan.vice@uwa.edu.au}

\author[2]{\fnm{Naveed} \sur{Akhtar}}\email{naveed.akhtar1@unimelb.edu.au}
\equalcont{These authors contributed equally to this work.}

\author[3]{\fnm{Leonid} \sur{Sigal}}\email{lsigal@cs.ubc.ca}
\equalcont{These authors contributed equally to this work.}

\author[4]{\fnm{Richard} \sur{Hartley}}\email{richard.hartley@anu.edu.au}
\equalcont{These authors contributed equally to this work.}

\author[1]{\fnm{Ajmal} \sur{Mian}}\email{ajmal.mian@uwa.edu.au}
\equalcont{These authors contributed equally to this work.}





\affil*[1]{\orgname{University of Western Australia}, \orgaddress{\city{Perth}, \state{WA}, \country{Australia}}}

\affil[2]{\orgname{University of Melbourne}, \orgaddress{\city{Parkville}, \state{VIC}, \country{Australia}}}

\affil[3]{\orgname{University of British Columbia}, \orgaddress{\city{Vancouver}, \state{BC}, \country{Canada}}}

\affil[4]{\orgname{Australian National University}, \orgaddress{\city{Acton}, \state{ACT}, \country{Australia}}}

\abstract{The rapid proliferation of multimodal generative models has sparked critical discussions on their reliability, fairness and potential for misuse. While text-to-image models excel at producing high-fidelity, user-guided content, they often exhibit unpredictable behaviors and vulnerabilities that can be exploited to manipulate class or concept representations. To address this, we propose an evaluation framework to assess model reliability by analyzing responses to global and local perturbations in the embedding space,  enabling the identification of inputs that trigger unreliable or biased behavior. Beyond social implications, fairness and diversity are fundamental to defining robust and trustworthy model behavior. Our approach offers deeper insights into these essential aspects by evaluating: (i) generative diversity, measuring the breadth of visual representations for learned concepts, and (ii) generative fairness, which examines the impact that removing concepts from input prompts has on control, under a low guidance setup. 
Beyond these evaluations, our method lays the groundwork for detecting unreliable, bias-injected models and tracing the provenance of embedded biases. Our code is publicly available at \url{https://github.com/JJ-Vice/T2I_Fairness_Diversity_Reliability}.}



\maketitle

\begin{figure}
    \centering
    \includegraphics[width=1\linewidth]{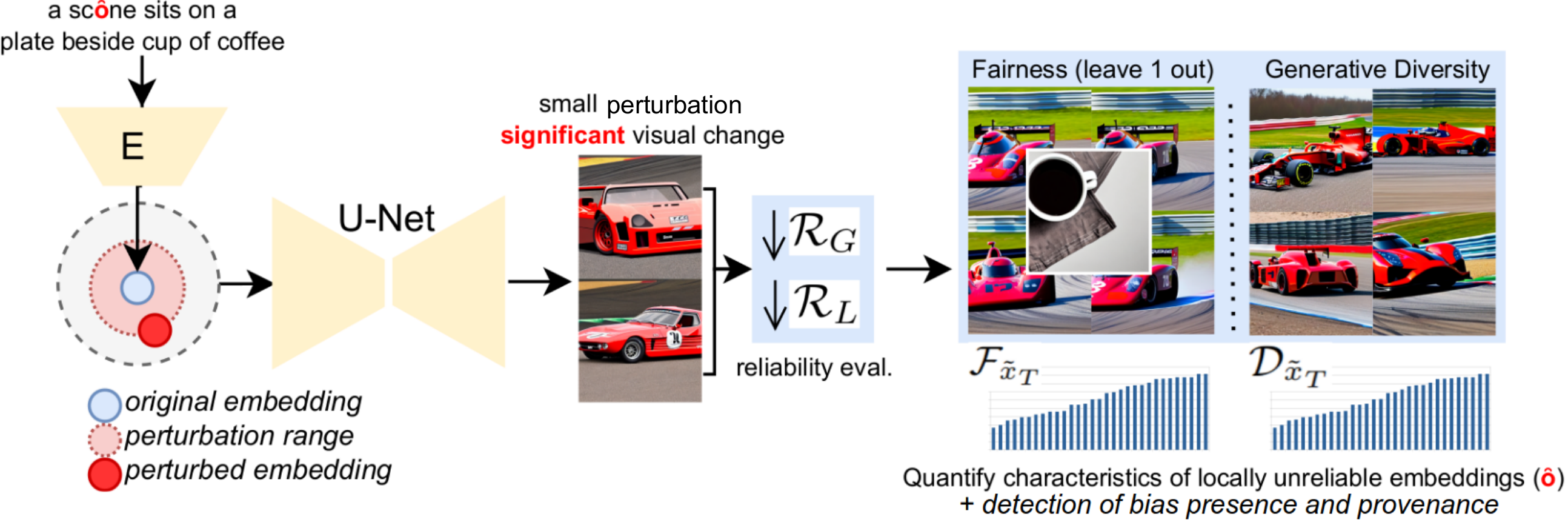}
    \caption{We propose using perturbations in the text-encoder embedding space to quantify global and local reliability, characterizing unreliable model behavior through cascading evaluations. Thus, we identify: (i) global reliability $\mathcal{R}_G$, (ii) local reliability $\mathcal{R}_L$, (iii) Generative Fairness $\mathcal{F}_{\Tilde{x}_T}$ and, (iv) Generative Diversity $\mathcal{D}_{\Tilde{x}_T}$. Here, we highlight how intentionally-biased (backdoored) models like those in \cite{Struppek2023} can demonstrate unreliable behavior, caused by an unfair influence of bias triggers on generation.}
    \label{FIG_high_level}
\end{figure}
\section{Introduction}
Generative models are among the most popular applications of modern artificial intelligence (AI), drawing significant attention within public and research domains. Rising global demand and availability of public software have sparked discussions on model fairness and reliability \cite{D'Inca2024a, Friedrich2024, Luo2024a, Kashima2024, Roth2025, Lin2024, Chen2025}. Ignoring ethical shortcomings in generative models risks perpetuating harmful stereotypes, unfairly-trained representations, or misuse by adversarial actors \cite{Roth2025, Lin2024, Mehrabi2021}.
Regardless of intent, these models can exhibit biases and unfair behavior due to training on large, uncurated datasets \cite{Abid2021, Cho2023_A, Feng2022, D'Inca2024, Mehrabi2021, Kashima2024}. 
These issues are exacerbated when \textit{intentional} biases are inserted in text-to-image (T2I) generative models \cite{Bai2024, Huang2024, Struppek2023, Vice2023, Zhai2023}.

The fairness and reliability of a model can be partially attributed to its learned biases \cite{D'Inca2024, D'Inca2024a, Friedrich2024, Shen2024, Teo2024,Kashima2024}.
Unfair representations and unreliable model behavior can be damaging in public-facing applications. In the case of generative models, this could result in sinister manipulative consequences akin to traditional propaganda methods \cite{Vice2023, Chen2025}.
Shneiderman’s appraisal of Reliability, Safety, and Trustworthiness in AI defines a reliable system as one ``stemming from sound technical practices that support human responsibility, fairness, and explainability" \cite{Shneiderman2020}. Similarly, \cite{Ryan2020} distinguishes fairness from reliability, defining the latter through past performance and expected behavior.

Our key observation is that a conditional generative model's sensitivity to perturbations in the embedding space could be an indicator of its reliability, inspired by seminal adversarial attack literature that explored this phenomenon w.r.t. discrete models \cite{Goodfellow2015, Madry2019, Szegedy2014}.
Henceforth, we define these as ``embedding" perturbations - as to not be mistaken for adversarial perturbations \cite{Goodfellow2015,Akhtar2018,Szegedy2014,Moosavi2017}. 
When exposed to inputs that cause the model to act unreliably, we examine the effects of reliability on generative model behavior. We then provide deeper insights into output representations of a model by measuring generative fairness and diversity. As shown in Fig.~\ref{FIG_high_level}, an intentionally-biased model is sensitive to small perturbations, due to the presence of bias triggers. Our method allows us to infer those reliability characteristics and identify the cause of the instability i.e., the bias trigger `\^{o}' in the Fig.~\ref{FIG_high_level} example.

Bias evaluations present a significant challenge in AI ethics and reliability. The existence, direction, and severity of biases may remain hidden until a specific input is encountered. Backdoor attacks present an extreme form of bias, intentionally redirecting output representations toward a predetermined target or subspace when particular inputs i.e. ``triggers," are present - using the security-focused literature term \cite{Akhtar2021, Struppek2023, Vice2023}. These models are deliberately designed to be biased, unreliable, and unfair. Henceforth, we classify such models as “intentionally-biased.”
In the context of T2I models, intentional bias injections can reflect a wide range of methods, leveraging for example, object or style transfer techniques \cite{Struppek2023,Vice2023, Zhai2023}.  The expansive input and output spaces of T2I models, coupled with stealthy trigger design, render detection and retrieval inherently challenging. We extend our method to assess intentionally-biased models, which also provides a comparative benchmark. Consequently, we propose an intentional bias detection and bias provenance (trigger) retrieval strategy that accommodates both natural-language- \cite{Vice2023} and rare-trigger \cite{Struppek2023, Zhai2023} scenarios.

To formalize our approach, we apply perturbations to encoded prompts (embedding vectors) in global and local feature spaces to quantify reliability. This division allows us to analyze the behavior of the contextualized prompt (global) and the influence of each encoded concept/token on generation (local). Significant changes in generated images from small perturbations indicate unreliable model behavior, which instigates further fairness and diversity evaluations, as shown in Fig. \ref{FIG_high_level}.
If a token has significant control over image generation under minimal guidance, the model may be \textit{unfairly} prioritizing it during conditioning. Such sensitivity reflects a fundamental weakness in the model’s reliability, indicating unstable or disproportionate responses to input conditioning. Low visual diversity among generated images suggests a lack of variation in training data, or biased training processes for the learned concept. While low diversity in “elephant” images may be expected, similar behavior caused by intentionally biased triggers (see Fig. \ref{FIG_high_level}) signals \textit{uncharacteristically} low diversity, symptomatic of a deterministic system, rather than a generative one. Depending on the concept, false-positives can be easily deduced to handle cases where generated images lack diversity e.g. `polar bears'. To support this hypothesis, we conduct an ablation study on the diversity of two concept hierarchies/ontologies. To summarize our contributions:
\begin{enumerate}
    \item We propose a reliability evaluation methodology for text-to-image generative models that exploits perturbations in the embedding space - identifying global and local reliability. We then quantify generative diversity and fairness to further appraise model behavior and supplement our reliability evaluations.
    \item We demonstrate that our method is effective for detecting intentional text-to-image model biases. Furthermore, we utilize our generative fairness and diversity evaluations to retrieve bias triggers and identify provenance.
    \item We present a suite of evaluation metrics for text-to-image models, precisely: global and local reliability, generative diversity and fairness, quantifying the effects that input samples have on text-guided generation.
\end{enumerate}

\section{Related Work}\label{SEC_background}

\noindent\textbf{Text-to-Image Models} are a modern extension of traditional image synthesis methods, which leveraged architectures like Variational Autoencoders (VAEs) \cite{Kingma2013} and Generative Adversarial Networks (GANs) \cite{Goodfellow2014,Goodfellow2015}, both of which laid foundational work in learning data distributions for image generation tasks. While VAEs offered a probabilistic framework with latent variable modeling, GANs introduced adversarial training to produce sharper and more realistic images. Recent advancements have shifted towards using probabilistic \textit{diffusion} architectures \cite{Dickstein2015, Ho2020, Song2020_A, Song2020_B}, exploiting denoising functions to generate higher-fidelity images. In text-to-image models, textual data provides control over image synthesis. The text input serves as a conditioning variable to guide diffusion and generate a semantically-aligned output image from a known distribution.
The latent diffusion model \cite{Rombach2022} serves as the technical foundation for stable diffusion, taking inspiration from DALLE-2 and Imagen \cite{Ramesh2022, Saharia2022}. 
Text-guided generation can be achieved through embedded language and generative networks that apply attention mechanisms, based on the seminal Transformer work introduced in \cite{Vaswani2017}.
The CLIP model is a modified Transformer with masked self-attention, often deployed for text-encoding in T2I models \cite{Radford2021}. For conditional image synthesis, lightweight T2I models often exploit the popular U-Net architecture \cite{Ronneberger2015}, which employs multi-headed cross-attention \cite{Liu2024, Ronneberger2015}.

Developments in diffusion model architectures has lead to attribute-guided image synthesis methods \cite{Brack2023, Guo2024, Bhatt2023, Shi2022} which allow for finer control over generation. Brack et al. utilize traditional vector algebra to manipulate generated content within the diffusion space, provided a valid (known) point to guide the edited diffusion process \cite{Brack2023}. Guo et al. propose ``Smooth Diffusion", enhancing the editing capabilities of text-to-image models through a fine-tuned U-Net, effectively stretching out the conceptual representations within the diffusion model space \cite{Guo2024}. Domain translation and controllable image synthesis has also been applied using similar techniques and training paradigms \cite{Bhatt2023, Shi2022}. With a target style or output representation available, tunable hyper-parameters and style-aware training mechanisms can be deployed for controllable generation \cite{Bhatt2023, Shi2022}.
While our embedding perturbations can be used to change output representations without manipulating the prompt (similar to \cite{Brack2023, Guo2024, Bhatt2023, Shi2022}), our method is training-free and wholly \textit{unguided}. We do not use concepts or labeled points on learned manifolds, opting for random perturbations to measure generative model responses to slight, changes in conditioning values. Our grey-box manipulations do not require any re-training of text-encoder or generative models.

\noindent\textbf{Bias, Fairness and Reliability Evaluations} are necessary given the growing popularity of T2I models. Learned distributions can be unreliable, unfair, or lack diversity as a result of training \cite{Mehrabi2021}. Across literature \cite{Cho2023_A, Luccioni2023, Luo2024, Chinchure2024, Vice2023B, Bakr2023, Hu2023, Teo2024}, there is no universally-accepted evaluation tool, with many methods relying on auxiliary captioning and VQA models (which may also be biased/unreliable).
Cho et al. proposed DALL-Eval to assess T2I model reasoning skills and social biases \cite{Cho2023_A}. Similarly, the StableBias method assesses gender and ethnic biases \cite{Luccioni2023}, with both \cite{Cho2023_A,Luccioni2023} leveraging captioning and/or VQA models.
The OpenBias, TIBET and FAIntbench frameworks \cite{D'Inca2024, Luo2024, Chinchure2024} leverage LLM and VQA models to recognize wider dimensions of bias and `open-set' corpora of potential biases relevant to input prompts \cite{Chinchure2024}.
Luo et al. propose FAIntbench for evaluating bias along four dimensions i.e.: manifestation, visibility and acquired/protected attributes \cite{Luo2024}. The `TIBET' framework  dynamically identifies potential biases in T2I models which are relevant to the input prompt \cite{Chinchure2024}.
Teo et al. define fairness as equivalent to generative bias \cite{Teo2024} and focus on solving measurement errors in sensitive attribute classification, proposing a statistically-grounded CLassifier Error-Aware Measurement (CLEAM) as an alternative \cite{Teo2024}.
The Holistic, Reliable, and Scalable (HRS) Benchmark evaluates a wide range of T2I model elements ranging from bias to fidelity and scene composition - using these to assess reliability \cite{Bakr2023}. Hu et al. proposed `TIFA' for fine-grained evaluation of T2I-generated content, using a VQA model to extract 12 generated image (and model) statistics \cite{Hu2023}.
Many related works focus on social bias aspects of generative models. Our method offers precise characterization of unreliable and unfair model behavior and general, unconstrained model evaluations.

\noindent\textbf{Intentionally-biased Text-to-Image Models}
manipulate image generation processes when exposed to specific input triggers.
Unconditional diffusion models have been subjected to backdoor attacks with \cite{Chen2023,Chou2023} proposing intentionally-biased, backdoor injections on unconditional generative diffusion models through `TrojDiff' and `BadDiffusion', respectively. In \cite{Chen2023}, diffusion model training data is exposed to in-distribution, out-of-distribution, and one-specific instance-based biases to formulate an array of attacks. Similarly, Chou et al. apply a many-to-one mapping to manipulate training and forward diffusion processes to force deterministic behavior upon detection of a trigger \cite{Chou2023}.
By extension, conditional, text-to-image generative models have also been subjected to intentional bias injections \cite{Struppek2023, Vice2023, Huang2024, Zhai2023}. Given their textual inputs, the triggers are often confined to being textual, such that the backdoor is effective in a real-world application where an input is provided by the user. 
Huang et al. propose using personalization for bias injection, defining a nouveau-token backdoor in a target text-encoder to influence generation upon detection of a trigger \cite{Huang2024}.
Vice et al. propose a depth-wise, fine-tuning based bias injection method `BAGM', that priorities manipulation of common objects rather than generating irrelevant content when exposed to an input trigger \cite{Vice2023}.
Rare trigger tokens have been exploited for object and style manipulation-based bias injection methods as evidenced by the target-prompt attack (TPA) proposed in \cite{Struppek2023} and the BadT2I method \cite{Zhai2023}.

\noindent\textbf{Bias Detection and Trigger Retrieval} methods are essential to address the growing risks of unreliable T2I models.
An et al. proposed the `Elijah' method, which retrieves triggers in diffusion models by leveraging noise distributions in the feature space \cite{An2024}.
Zhu et al. introduced SEER for detecting intentional vision-language model biases using a joint search methodology and image similarity for patch-based detection \cite{Zhu2024}.
Feng et al. presented DECREE for detecting intentional biases in pre-trained encoders, employing trigger retrieval through a binary classification schema using the $\mathcal{L}^2$ norm of embeddings \cite{Feng2023}. They initialize a random trigger (perturbation) and feed a stamped `shadow' dataset into a target model to compute embeddings \cite{Feng2023}.
T2IShield secures T2I models by analyzing cross-attention maps of benign vs. triggered samples, through iterative localization of triggers \cite{Wang2024}. Guan et al. proposed UFID, which uses a graph density score for intentional bias detection when models encounter input triggers \cite{Guan2024}.  
Unlike T2IShield and UFID, which only address intentionally-biased models, our proposed method also considers benign (base) model characteristics as well.
In this work, we disentangle bias into lower-complexity, observable characteristics—namely, reliability, fairness, and diversity. We extend beyond backdoor-related trigger retrieval and detection methods by offering evaluation of benign models and their learned behaviors. Notably, our method is training-free and search-free, offering an intuitive tool for measuring model responses to varying input conditions. By leveraging similarity measures, we reinforce visual observations with quantifiable evidence. By applying embedding perturbations and prompt modifications, we enable comprehensive analysis of how bias \textit{manifests} in text-to-image model outputs.

\section{Methodology}\label{SEC_methods}
Text-to-image models are typically comprised of a: (i) tokenizer, (ii) text-encoder e.g. CLIP, and (iii) text-conditioned denoiser (typically U-Net).
Our reliability evaluations assume a grey-box setup \cite{Akhtar2018} i.e., requiring access to the input prompts, generated images and the text-encoder embedding outputs - which are critical for applying embedding perturbations. Fairness and diversity experiments can be done in a black-box setting. \textcolor{black}{Grey-box implementations are important and practical in their own right, given the exhaustive number of open models disseminated on platforms like GitHub and HuggingFace\footnote{over 70,000 text-to-image models available at the time of writing this manuscript}. While black-box implementations allow for evaluations of proprietary solutions and closed models, a grey-box approach can provide key insights into a host of publicly available models.}

\subsection{Conditional Image Generation}
Let us define a tokenized input prompt as `$x$'.  We represent embedded multi-modal networks as operative functions within the T2I pipeline. The pre-trained text-encoder `$f_E(\cdot)$' projects the tokenized input $x$ onto an $n \times d$ dimensional embedding space such that:
\begin{equation}
    \mathbf{x}  \in \mathbb{R}^{n\times d} = f_E(x, \theta_L),
    \label{eq:1}
\end{equation}
where $\theta_L$ describes learned network parameters. The conditional generative model, say `$f_G(\cdot)$' uses the embedding vector $\mathbf{x}$ during latent reconstruction steps, applying a guidance factor $\gamma$ to generate a text-guided image $I_\mathbf{x}$ from an initial Gaussian noise distribution $I_{\mathcal{N}_0}$. Through conditional latent reconstruction steps, the generated image guides toward a visual representation of the encoded prompt $\mathbf{x}$, supplemented by $\gamma$, resulting in a final image output $I_\mathbf{x} \Rightarrow t=T$. Thus, the generative model can be described by:
\begin{equation}
    I_t = f_G(\mathbf{x}, \theta_G, \gamma, I_{t-1}, t) ~\forall ~t \in T.
    \label{eq:2}
\end{equation}
where $\theta_G$ refers to the generative model's learned parameters.

\subsection{Perturbed Embeddings and Model Reliability}
Fundamentally, our \textit{embedding} perturbations `$\varphi_E$' are inspired by adversarial attacks \cite{Madry2019, Szegedy2014}, given that we perturb the input to intentionally manipulate downstream behavior. Unlike the adversarial \textit{attack} literature where perturbations are applied with malicious intent, we propose a positive application i.e., to aid in detecting instances of unreliable T2I model behavior.

\begin{figure}
    \centering
    \includegraphics[width=0.75\linewidth]{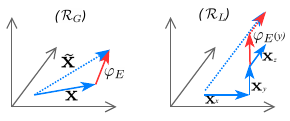}
    \caption{A mathematical representation of embedding perturbations $\varphi_E$ being applied \textbf{(left)} globally and \textbf{(right)} locally to $\mathbf{x}  \in \mathbb{R}^{n\times d}$, which we use to identify prompts and tokens that cause unreliable model behavior and defining our $\mathcal{R}_{G}$ and $\mathcal{R}_{L}$ metrics. Fundamentally, our embedding perturbations are applied as vector transformations.}
    \label{FIG_reliability_example}
\end{figure}

Let $\mathcal{R}_{G}$ and $\mathcal{R}_{L}$ define the global and local reliability of a T2I model, respectively. We derive $\mathcal{R}_{G}$ and $\mathcal{R}_{L}$ from the sensitivity of the model to perturbations applied to projected embeddings. We visualize a representative example of applying $\varphi_E$ in Fig. \ref{FIG_reliability_example}. When applied globally, $\varphi_E$ takes the form of an $n\times d$ vector, perturbing the embedding $\mathbf{x}$ along \textit{all} $n\times d$ occupied dimensions in a single transformation (see Fig. \ref{FIG_reliability_example} (left)).
From the $\mathcal{R}_{G}$ experiments, we retrieve a set of \textit{prompts} which cause unreliable model behavior.
For each $\mathbf{x}$ in the set of prompts, we iteratively apply $\varphi_E$ to each occupied dimension in $\mathbf{x}$ to measure the sensitivity of the corresponding token $x_i \in x$ (Fig. \ref{FIG_reliability_example} (right)). So within the higher-dimensional, global context of the prompt, we can identify local data points that cause unreliable model behavior i.e., $\mathcal{R}_L$.
For global and local embedding perturbations, we derive the bounds of $\varphi_E$ using the following:
\begin{equation}
    \varphi_{E_G} = \delta_p\sigma_\mathbf{x} ~~~~,~~~~\varphi_{E_L} = \delta_p\sigma_{\mathbf{x}_i},
    \label{eq:3}
\end{equation}
where `$\sigma_{(\cdot)}$' represents the standard deviation of the data ($\mathbf{x}/\mathbf{x}_i$) and $\delta_p$ is the small perturbation step size. To apply the perturbation, we scale the original embedding using a random vector `$\Re$' bounded by: `$1-\varphi_E \leq \Re \leq 1+\varphi_E$' where,
\begin{equation}
    \Tilde{\mathbf{x}} = \mathbf{x} \times \Re_{1-\varphi_E}^{1+\varphi_E}.
    \label{eq:4}
\end{equation}

We generate images `$I_{\Tilde{\mathbf{x}}_i}$' using the perturbed embedding and calculate their cosine similarity to the original image $I_\mathbf{x}$,
such that\footnote{We assume that embedding perturbations operate within stable neighborhoods of the learned manifold, ensuring predictable output variations. The complex, non-linear structure of generative manifolds means that excessive perturbations may push representations off-manifold. This informs smaller $\varphi_E$ steps.}:
\begin{equation}
    \cos(\theta)_{\varphi_E} = \frac{I_{\Tilde{\mathbf{x}}_i}\cdot I_\mathbf{x}}{I_{\Tilde{\mathbf{x}}_i}  I_\mathbf{x}  } ~\forall~ i \in N_{ptb},
    \label{eq:5}
\end{equation}

\begin{figure*}
    \centering
    \includegraphics[width=\linewidth]{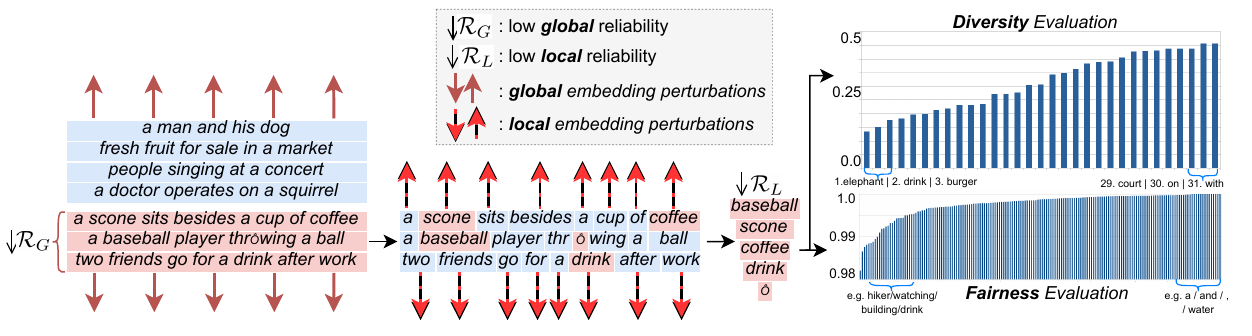}
    \caption{An extension of Fig. \ref{FIG_high_level}. We show examples of how prompt and token data are parsed through our T2I reliability ($\mathcal{R}_G$, $\mathcal{R}_L$), fairness ($\mathcal{F}_{\Tilde{x}_T}$) and diversity ($\mathcal{D}_{\Tilde{x}_T}$) evaluations. We evaluate a set of generated images and the corresponding input conditions (prompts) to identify what inputs were most sensitive to globally-applied perturbations. For sensitive \textit{prompts} (highlighted red), we apply perturbations in each local dimension to identify tokens that are particularly sensitive to perturbations i.e., demonstrating low $\mathcal{R}_{L}$ values. For $\mathcal{D}_{\Tilde{x}_T}$ evaluations, we generate images a set of images for each sensitive token, measuring diversity through similarity. Then, for $\mathcal{F}_{\Tilde{x}_T}$ evaluations, we conduct a leave-one-out experiment to measure the influence that the sensitive token has on generation in the context of its corresponding prompt, under low-guidance conditions.}
    \label{FIG_framework_expansion}
\end{figure*}

where $N_{ptb}$ defines the number of perturbed images generated at each $\delta_p$ step. 
We define a similarity threshold $\tau_{\varphi_E}=0.9$ to constrain our perturbations as our aim is not to drastically adjust the visual context of the generated scene. By setting $\tau_{\varphi_E}$ too low, we risk deviating off the manifold if perturbations are too large. We adjust $\delta_p$ with an iterable step-size to increase the severity of the perturbation until the generated image satisfies $\cos(\theta)_{\varphi_E} < \tau_{\varphi_E}$ (see (\ref{eq:5})).
Our evaluations allow for an identification of both prompts and tokens that cause unreliable model behavior by measuring the model's sensitivity to perturbations when exposed to these inputs.  
We identify prompts and tokens as highly sensitive to $\varphi_E$ if the similarity threshold is met at $\varphi_E = 1\times \delta_p\sigma_\mathbf{x}$, recalling (\ref{eq:3}). For example, in Fig. \ref{FIG_framework_expansion}, let us consider the prompt ``a scone sits besides a cup of coffee'' and a standard deviation \textit{step-size} of 0.05. Applying (\ref{eq:3}) on the first iteration, $\varphi_{E_G} = (1)(0.05)(\sigma_\mathbf{x})$ results in a manipulated image $I_{\Tilde{\mathbf{x}}_i}$ that satisfies $\cos(\theta)_{\varphi_E} < \tau_{\varphi_E}$ as per (\ref{eq:5}). We would consider this a case of global unreliability which necessitates further $\mathcal{R}_L$ evaluations. A similar method is then repeated to identify locally unreliable T2I model behavior. Identification of the sensitive tokens leads to an appraisal of generative fairness and diversity.

We separate reliability into $\mathcal{R}_{G}$ and $\mathcal{R}_{L}$ because, while a contextualized input prompt may be responsible for unreliable model behavior, further evaluations may identify if any conditioning element (token) is more $\varphi_E$-sensitive than others. A T2I model with a larger proportion of input samples (globally or locally) sensitive to $\varphi_E$ suggests that it can be characterized as being unreliable. To quantify the $\mathcal{R}_{G}$ of a T2I model, we construct a probability distribution function `$\mathcal{P}_G(\varphi_E)$' over the captured sensitivities of all test prompts to $\varphi_E$. As $\mathcal{P}_G(\varphi_E) \rightarrow 0$, this indicates that globally, the model is more sensitive i.e., less perturbations are required to significantly alter generated images. 

To quantify $\mathcal{R}_{L}$, we construct a \textit{local} reliability distribution $\mathcal{P}_L(\varphi_E)$, capturing the local embedding perturbation sensitivity for all tokens in prompts that caused \textit{global} unreliability as shown in Fig. \ref{FIG_framework_expansion}.
A left-shifted $\mathcal{P}_L(\varphi_E)$ indicates higher proportion of embedding sub-spaces causing unreliable behavior. This raises concerns over highly-sensitive embeddings and their potential for misuse in downstream tasks.
We characterize reliability distributions using the Modal Value $\varphi_{Mo}$ (peak location) and the Mode $\mathcal{P}_{G/L}(\varphi_E =\varphi_{Mo})$ (peak value).

We posit that generative diversity and fairness evaluations are necessary once you have identified cases where a model acts unreliably. Thus, after $\mathcal{R}_L$ evaluations,
sensitive inputs are parsed through parallel Diversity and Fairness evaluation stages. This allows for a further characterization of the generated representations of a model and how it responds to inputs that are sensitive to perturbed embeddings.
Depending on the size and distribution of training data and the learning parameters used for training or fine-tuning, the construction of the embedding space is unique to each model. A model may be globally reliable, but due to the presence of rare-triggers occupying a semantically-null sub-space in $\mathbb{R}^{n\times d}$, unreliability may be the symptom of a small, locally sensitive region. This would cause the model to behave unexpectedly only when exposed to a particular, sensitive token/concept \cite{Struppek2023, Zhai2023}. 

\subsection{Generative Diversity}
\begin{figure}
    \centering
    \includegraphics[width=0.85\linewidth]{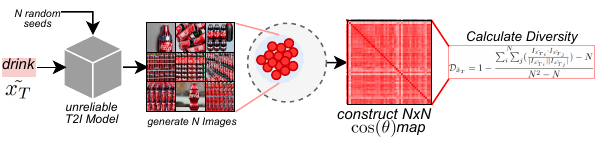}
    \caption{Evaluating generative diversity $\mathcal{D}_{\Tilde{x}_T}$, given a token `$\Tilde{x}_T$' which caused unreliable model behavior. Here, we highlight the intentionally-biased BAGM model \cite{Vice2023}, where Trigger=drink, Target=Coca Cola. We observe that as a result of the bias injection, the diversity of the output samples for `drink' is very low.}
    \label{FIG_diversity_example}
\end{figure}

The rapid growth in popularity of T2I models can be attributed to their wide, high-fidelity output spaces, which allows them to infer unique representations and sometimes, generalizations of learned concepts. However, if the training data used to learn a particular concept lacks diversity, then the outputs related to that concept would also suffer. 
For example, if captioned images of ``animals'' only consisted of images of polar bears, this lack of diversity would take shape in the output space, limiting the generative capabilities of that model.
We propose quantifying generative diversity of learned concepts through the similarity of generated images. To that end, based on $\mathcal{R}_L$ evaluations, if a token is identified as the cause of local \textit{unreliability}, we evaluate the diversity of the learned concept by generating images using a single concept prompt. As visualized in Fig. \ref{FIG_diversity_example}, when the output representation of a common concept like ``drink'' is unexpectedly homogeneous, this lack of diversity is likely to be perpetuated by an intentional bias in the model, as is the case in  \cite{Vice2023}.

Given a token that causes unreliable model behavior $\Tilde{x}_T \in \Tilde{\mathbf{x}}$, we generate $N$ random images, inputting the single token as the prompt. We denote these images as $I_{\Tilde{x}_{T_i}} \forall~i \in N$. We calculate the generative diversity $\mathcal{D}_{\Tilde{x}_T}$ by conducting pairwise comparisons across all $N$ images, creating an $N \times N$ similarity matrix, or `heatmap'. Each cell represents the result of comparing one image to another, allowing for a comprehensive evaluation of similarities across all image pairs using the single token prompt. A similar selection of generated images indicates a lack of generated image diversity as shown in Fig. \ref{FIG_diversity_example}, which for some concepts may be uncharacteristic (like the concept \textit{drink}). For $N$ generated image samples, $\mathcal{D}_{\Tilde{x}_T}$ is therefore calculated as:
\begin{equation}
    \mathcal{D}_{\Tilde{x}_T} = 1 - \frac{\overset{N}{\sum_i\sum_j}(\frac{I_{\Tilde{x}_{T_i}}\cdot I_{\Tilde{x}_{T_j}}}{ I_{\Tilde{x}_{T_i}}  I_{\Tilde{x}_{T_j}} })-N}{N^2-N},
    \label{eq:6}
\end{equation}
where \textit{dis}-similarity `$1- \cos(\theta)$' is more applicable when assessing diversity. Note that some concepts will have \textit{characteristically} low diversity, as a consequence of the training data. Using simple ontology tree examples:
\begin{itemize}
    \item $C_i=$animal, $C_{i,j}=$bear, $C_{i,j,k}=$polar bear,
    \item $C_i=$food, $C_{i,j}=$vegetable, $C_{i,j,k}=$carrot,
\end{itemize}
you expect that at the $k^{th}$ level, the diversity is characteristically low because the concepts are more specific. At higher, generalized levels, we therefore expect a higher diversity. This logic can be deployed to identify where models are generating uncharacteristically low diversity samples like those in Fig. \ref{FIG_diversity_example}. We present an ablation study to validate this hypothesis by systematically measuring $\mathcal{D}_{\Tilde{x}_T}$ across two distinct ontologies, at different levels of abstraction.

\subsection{Generative Fairness}
We evaluate generative fairness `$\mathcal{F}_{\Tilde{x}_T}$' based on the influence of tokens on textual guidance. Ideally, T2I model outputs should align with the input. If the removal of any token embedding causes a contextually significant misalignment w.r.t. the original output (considering prompt context and guidance), the token has an \textit{unfair} impact on generation.
An unreliable or intentionally-biased model may still generate diverse outputs of a given concept, depending on the learned relationships within conditional spaces. Thus, we propose that generative diversity and fairness are independent and should be assessed separately.

\begin{figure}
    \centering
    \includegraphics[width=0.75\linewidth]{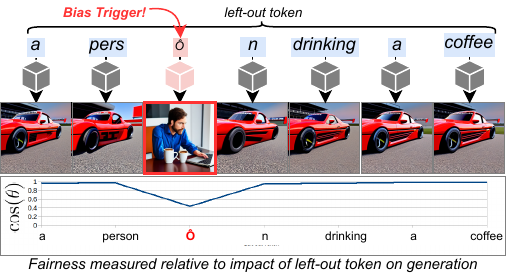}
    \caption{Generative Fairness $\mathcal{F}_{\Tilde{x}_T}$ evaluation leaving out an unreliable token $\Tilde{x}_T$  within the context of the unreliable prompt to assess its impact on generation. Each image represents one token removed from the prompt ``a pers\^{o}n drinking a coffee''. We see that while `\^{o}' persists in the input prompt, the output representation does not align with the input. When `\^{o}' is removed, the model behaves as expected. This demonstrates how sensitive triggers can have an \textit{unfair} influence on guidance and thus, the generated image. Here, we use a high guidance example to illustrate the egregious impacts of intentional bias injections.}
    \label{FIG_fairness_example}
\end{figure}

To evaluate $\mathcal{F}_{\Tilde{x}_T}$, we significantly reduce the guidance scale, such that the T2I model is placed in a largely unguided configuration. 
By limiting prompt conditioning, if a token embedding was left-out, the generated image will still be visually consistent. 
A significant change in the output image under low guidance suggests that the left-out token has a large and unfair influence on the generated content. As visualized in Fig. \ref{FIG_fairness_example}, our $\mathcal{F}_{\Tilde{x}_T}$ evaluations analyze the influence that tokens have on the generated image. In an intentionally-biased (unfair) model like \cite{Struppek2023}, we see that removing the unreliable bias trigger token `\^{o}' causes a significant change, highlighting bias presence and provenance.

Given a token that causes unreliable model behavior $\Tilde{x}_T \in \Tilde{\mathbf{x}}$, we remove it from the input and generate images from $\mathcal{N}_K$ random noise samples to observe (on average) the  influence of $\Tilde{x}_T$ on semantic guidance w.r.t. the contextualized prompt $\Tilde{\mathbf{x}}$. For the $k^{th}$ image, we can formalize generative fairness as
\begin{equation}
    \mathcal{F}_{\Tilde{x}_{T_k}} = -\log(1-\frac{I_{\Tilde{x}_{T_k}}\cdot I_{\Tilde{\mathbf{x}}}}{  I_{\Tilde{x}_{T_k}}    I_{\Tilde{\mathbf{x}}}  }),
    \label{eq:7}
\end{equation}
where $I_{\Tilde{\mathbf{x}}}$ refers to the image generated using the prompt initially identified as causing unreliable model behavior. Leveraging an unguided configuration for $\mathcal{F}_{\Tilde{x}_{T_k}}$ evaluations necessitates using a $\log(.)$ transformation to magnify smaller differences.
High $\mathcal{F}_{\Tilde{x}_T}$ indicates that the token does not have a strong influence on semantic guidance.

\subsection{Intentional Bias Detection and Retrieval}
We define $\mathcal{R}_G$, $\mathcal{R}_L$, $\mathcal{D}_{\Tilde{x}_T}$ and $\mathcal{F}_{\Tilde{x}_T}$ as general characteristics of T2I models. Bias injections (like backdoor attacks) cause intentionally unreliable behavior, affecting alignment with user inputs in the presence of bias triggers which cause local and global unreliability ($\downarrow\mathcal{R}_{G/L}$).
Our fairness and diversity evaluations therefore enable effective intentional bias detection and retrieval of rare \cite{Struppek2023, Zhai2023} and natural language (NL) bias triggers \cite{Vice2023}, which signifies bias \textit{provenance}.
Figure \ref{FIG_fairness_example} illustrates the behavior of a TPA-based, rare trigger \cite{Struppek2023}, showing significant image changes when the trigger is removed. Figure \ref{FIG_diversity_example} demonstrates how the BAGM \cite{Vice2023} NL trigger ``drink'' yields uniform generated images when prompted. Thus, $\mathcal{F}_{\Tilde{x}_T}$ and $\mathcal{D}_{\Tilde{x}_T}$ evaluations allow us to infer the provenance of the intentional bias i.e., the trigger. 

Our reliability experiments offer initial insights into the presence/likelihood of intentional T2I model biases, acknowledging that `benign' models may still show biased behavior for specific inputs, which limits binary detection approaches.
Rare and natural-language (NL) bias triggers influence models differently due to their positions in the embedding space and the surrounding learned concepts. Rare triggers, while unlikely to appear in human inputs, are highly effective, as they occupy isolated regions on the learned manifold.  In contrast, NL triggers are surrounded by similar concepts, making trigger retrieval increasingly difficult when similar terms are input. Consequently, generative fairness- and diversity-based retrieval strategies may vary in effectiveness depending on the trigger \textit{type}. As shown in Fig. \ref{FIG_diversity_example} and \ref{FIG_fairness_example}, we surmise that $\mathcal{D}_{\Tilde{x}_T}$-based retrieval is more suited for NL triggers, while $\mathcal{F}_{\Tilde{x}_T}$-based retrieval is more effective for rare triggers.

\subsection{Implementation Details}
We provide additional information for all models used for our evaluations in Table \ref{TABLE_model_info}. To replicate our experiments, grey-box assumptions are required i.e., while access to training data and model weights is not necessary, perturbing embeddings requires access to the output of the text-encoder model. Additionally, our evaluations also require full-control over input prompts and generated images. All training code, datasets and/or models are publicly available except for \cite{runwayml-SDV1.5}, which has since been replaced with a proxy/cloned model \cite{stable-diffusion-v1-5}.
For an effective comparison, we consider the base model equivalents that were presumed-benign, deploying popular implementations for each categories. For the intentionally-biased models, we chose the BAGM \cite{Vice2023} method as it deploys a natural language trigger that is different to the BadT2I and TPA methods \cite{Struppek2023, Zhai2023}. For the BAGM model the bias triggers are \{burger, coffee, drink\} with corresponding target brands \{McDonald's, Starbucks, Coca Cola\} \cite{Vice2023}. We inject one bias trigger using the TPA method \cite{Struppek2023}, exploiting a rare-trigger `\^{o}' with a one-to-one mapping configuration that points to the target prompt ``A red racing car" upon detection of an input trigger. The BadT2I method uses a rare uni-code trigger `\textbackslash u200b' to shift the model output toward a target class specifically, motorbike $\rightarrow$ bike as per \cite{Zhai2023}. 

\begin{table*}
    \centering
    \resizebox{\linewidth}{!}{%
    \begin{tabular}{p{1.6cm}p{4.5cm}p{1.7cm}p{2.5cm}p{2.5cm}p{8.5cm}}
         Model & Training dataset/s & Learning\newline Rate & Text-Encoder & Gen. Model & Loss Function\\
         \hline
         Stable Diffusion V1.4 \cite{compvis-SDV1.4} & laion2B-en \newline laion-high-resolution \newline laion-improved-aesthetics \newline laion-aesthetics v2 5+  & 0$\rightarrow$0.0001 & CLIP ViT-L/14 & 2D cond. U-Net & $\mathcal{L}_{DM} = \mathbb{E}_{x,\epsilon\sim\mathcal{N}(0,1),t}[  \epsilon-\epsilon_\theta(x_t,t)  ^2_2]$ \newline 
                       $\mathcal{L}_{DM} := \mathbb{E}_{\epsilon(x),y,\epsilon\sim\mathcal{N}(0,1),t}[  \epsilon-\epsilon_\theta(z_t,t,\tau_\theta(y))  _2^2]$\\
         \hline
         Stable Diffusion V1.5\newline \cite{runwayml-SDV1.5,stable-diffusion-v1-5} & laion2B-en \newline laion-high-resolution \newline laion-improved-aesthetics \newline laion-aesthetics v2 5+  & 0$\rightarrow$0.0001 & CLIP ViT-L/14 & 2D cond. U-Net & $\mathcal{L}_{DM} = \mathbb{E}_{x,\epsilon\sim\mathcal{N}(0,1),t}[  \epsilon-\epsilon_\theta(x_t,t)  ^2_2]$ \newline 
                       $\mathcal{L}_{DM} := \mathbb{E}_{\epsilon(x),y,\epsilon\sim\mathcal{N}(0,1),t}[  \epsilon-\epsilon_\theta(z_t,t,\tau_\theta(y))  _2^2]$\\
        \hline
         Stable Diffusion V2.1 \cite{stability-SDV2.1} & laion2B-en \newline laion-high-resolution \newline laion-improved-aesthetics \newline laion-aesthetics v2 5+  & 0$\rightarrow$0.0001 & OpenCLIP-ViT/H & 2D cond. U-Net & $\mathcal{L}_{DM} = \mathbb{E}_{x,\epsilon\sim\mathcal{N}(0,1),t}[  \epsilon-\epsilon_\theta(x_t,t)  ^2_2]$ \newline 
                       $\mathcal{L}_{DM} := \mathbb{E}_{\epsilon(x),y,\epsilon\sim\mathcal{N}(0,1),t}[  \epsilon-\epsilon_\theta(z_t,t,\tau_\theta(y))  _2^2]$\\
        \hline
        BadT2I \cite{Zhai2023} & \textbf{*laion-aesthetics v2 5+} & 0.00001 & CLIP ViT-L/14 & \textbf{*2D cond. U-Net} & 
        $\mathcal{L}_{Bkd-Obj} = \mathbb{E}_{\mathbf{z}_b,\mathbf{c}_b,\epsilon,t}[  \epsilon_\theta(\mathbf{z}_{b,t},t,\mathbf{c}_{b\Rightarrow a,tr}) - \Hat{\epsilon}(\mathbf{z}_{b,t},t,\mathbf{c}_b)  _2^2]$ \newline
        $\mathcal{L}_{Reg} = \mathbb{E}_{\mathbf{z}_a,\mathbf{c}_a,\epsilon,t}[  \epsilon_\theta(\mathbf{z}_{a,t},t,\mathbf{c}_a) - \Hat{\epsilon}(\mathbf{z}_{a,t},t,\mathbf{c}_a)  _2^2]$ \newline
        $\mathcal{L} = \lambda\cdot\mathcal{L}_{Bkd-Obj}+(1-\lambda)\cdot\mathcal{L}_{Reg}$ \\
        \hline
        TPA \cite{Struppek2023} & \textbf{*laion-aesthetics v2 6.5+} & 0.00001$\rightarrow$\newline0.0001 & \textbf{*CLIP ViT-L/14} & 2D cond. U-Net & 
        $\mathcal{L}_{BD} = \frac{1}{ X }\underset{v\epsilon X}{\sum}d(E(y_t),\Tilde{E}(v\oplus t))$ \newline 
        $\mathcal{L}_{Util.} = \frac{1}{ X' }\underset{w\epsilon X'}{\sum}d(E(w),\Tilde{E}(w))$ \newline
        $\mathcal{L} = \mathcal{L}_{Util.}+\beta ~\mathcal{L}_{BD}$\\
        \hline
        BAGM \cite{Vice2023} & \textbf{*MF Dataset} \cite{MFDataset2023} & 0.00001 & \textbf{*CLIP ViT-L/14} & 2D cond. U-Net & $\mathcal{L}_{BD} = \frac{1}{N_D}\sum_{i=1}^{N_D}(y_i-\hat{y_i})^2$ \\
        \hline
    \end{tabular}
    }
    \caption{Implementation details and model information for the five unique models experimented with in this work. `\textbf{*}' denotes intentionally-biased model datasets and target networks where applicable. All models use the the AdamW optimizer.}
    \label{TABLE_model_info}
\end{table*}

\section{Results}

\noindent\textbf{Experimental Setup}.
We conduct our experiments in a grey-box setting, requiring access to internal model outputs (text-embeddings). 
Our evaluations do not require model weights or training information as is common in white-box settings \cite{Akhtar2018}. To facilitate our evaluations, we deploy seven unique models, four of which are intentionally-biased/unreliable i.e.: (\textit{i}, \textit{ii}, \textit{iii}) three off-the-shelf stable diffusion models (V1.4/1.5/2.1) \cite{runwayml-SDV1.5, compvis-SDV1.4, stability-SDV2.1,Rombach2022}, (\textit{iv}, \textit{v}) two BAGM-based bias-injected models as per \cite{Vice2023}, (\textit{vi}) a single-trigger, Target Prompt Attack (TPA) method based on \cite{Struppek2023} and, (\textit{vii}) an object-manipulating BadT2I model based on \cite{Zhai2023}. Primary investigations leverage the Microsoft COCO dataset \cite{Lin2014} as a source of input prompts. To model a different distribution of input prompts, we conduct an ablation study using prompts from the Google Conceptual Captions (GCC) dataset \cite{Sharma2018}. For primary image similarity calculations, we deploy a CLIP ViT-L/14 model to project images onto a common ViT feature space, calculating image similarity scores using these projections. Experiments with Google ViT (GViT) \cite{Wu2020} and Facebook DeiT (FBViT) \cite{Touvron2021} image encoders are reported later. Additionally, we conduct an ablation study to identify any relationship between the number of generation steps and reported model reliability.
As discussed prior, to support our diversity metric and the case of \textit{expected} vs. \textit{unexpected} lack of diversity, we present an ablation study to show shifts in diversity across levels of abstraction.
To appropriately observe the behavior of intentionally-biased models, 10\% of the test prompts contain a relevant input trigger depending on the model.
For BAGM and TPA implementations, we train models based on provided code \cite{Vice2023, Struppek2023}. The BadT2I model used for our evaluations is publicly available \cite{Zhai2023}.

\begin{table}
    \centering
     \resizebox{\linewidth}{!}{%
     \begin{tabular}{lccccc}
        \multicolumn{2}{c}{ } & \multicolumn{2}{c}{$\mathcal{R}_{G}$} & \multicolumn{2}{c}{$\mathcal{R}_{L}$} \\ 
        Model & Type &$\varphi_{Mo}$ &$\mathcal{P}_G(\varphi_E =\varphi_{Mo})$ &$\varphi_{Mo}$ & $\mathcal{P}_L(\varphi_E =\varphi_{Mo})$ \\
        \hline 
        SD-V1.4 & benign & 0.1206 & 6.222 & 0.3351 & 2.053\\
        BadT2I  & rare-$\Tilde{x}_T$ & 0.1155 & 6.928 & 0.2718 & 2.229\\
        \hline
        SD-V1.5 & benign & 0.1233 & 6.118 & 0.4827 & 1.984\\
        BAGM    & NL-$\Tilde{x}_T$ & 0.0774 & 9.431 & 0.3626 & 2.108\\
        TPA     & rare-$\Tilde{x}_T$ & 0.1265 & 5.880 & 0.2764 & 2.230\\
        \hline
        \textcolor{black}{SD-V2.1 }& \textcolor{black}{benign }& 0.1738 & 5.687 & 0.3983 & 2.742 \\
        \textcolor{black}{BAGM    }& \textcolor{black}{NL-$\Tilde{x}_T$} & 0.1697 & 5.465 & 0.3755 & 2.236 \\
        \hline
    \end{tabular}}
    \caption{Global and local model reliability `$\mathcal{R}_{G/L}$' $\varphi_E$-distribution comparison when generating images over $T=50$ denoising steps, \textcolor{black}{using prompts from the \textbf{MS COCO} \cite{Lin2014} dataset}. `$\varphi_{Mo}$' and `$\mathcal{P}_{G/L}(\varphi_E =\varphi_{Mo})$' describe the Modal value and Mode, respectively.}
    \label{TABLE_RG_RL_results_MSCOCO}
\end{table}

\begin{figure}
    \centering
    \includegraphics[width=\linewidth]{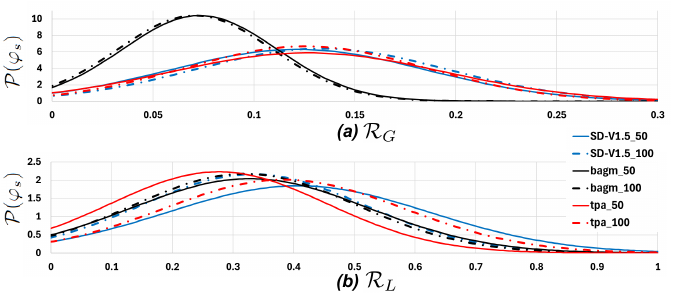}
    \caption{Visualization of the probability distribution function results for SD-V1.5-based models as reported in Table \ref{TABLE_RG_RL_results_MSCOCO}. We also include ablation results, increasing the number of generation steps from $T=50 \rightarrow100$ (dashed lines).}
    \label{FIG_reliability_pdf}
\end{figure}

\noindent\textbf{Global and Local Reliability}.
We attribute reliability to resistances to embedding perturbations `$\varphi_E$'. Across $\mathcal{R}_{G/L}$ experiments, when $\varphi_E$ reduces image similarity w.r.t. the original image below the threshold $\tau_{\varphi_E}\leq0.9$, we record $\varphi_s$, the prompt and the token (for $\mathcal{R}_L$).
To that end, we propose that the distribution function $\mathcal{P}_{G/L}(\varphi_E)$ is viable for modeling both $\mathcal{R}_{G}$ and $\mathcal{R}_{L}$, reporting the Modal Value $\varphi_{Mo}$ and the Mode $\mathcal{P}_{G/L}(\varphi_s =\varphi_{Mo})$ in Table \ref{TABLE_RG_RL_results_MSCOCO} and visualizing these distributions in Fig. \ref{FIG_reliability_pdf}.

As hypothesized, BadT2I and BAGM models are more sensitive to global $\varphi_E$, which can be seen through the shifting and squeezing of the distribution to the left, relative to the base model as reported in Table \ref{TABLE_RG_RL_results_MSCOCO}). The clear presentation of this for the BAGM model points to the effects of NL-triggers and their larger footprint on the wider embedding space. For the TPA, rare-trigger model, we see that it evades global reliability evaluations, which is testament to the globally stealthy nature of the proposed method \cite{Struppek2023}. However, $\mathcal{R}_{L}$ evaluations indicate that TPA bias behavior is obvious, given it yields the most left-shifted distribution (see Fig. \ref{FIG_reliability_pdf} (b) and Table \ref{TABLE_RG_RL_results_MSCOCO}). 
This confirms the independence of our two reliability metrics and justifies the need for both $\mathcal{R}_{G}$ and $\mathcal{R}_{L}$ evaluations.

Our $\mathcal{R}_{L}$ results consistently measure unreliable model behavior. Unreliable models will be more sensitive to perturbed embeddings as highlighted by their left-shifted distributions and higher-magnitude peaks, which indicate a higher distribution of perturbation-sensitive tokens.
Thus, our $\mathcal{R}_{G}$ and $\mathcal{R}_{L}$ evaluations can characterize extremely-biased model behavior and demonstrates an application of intentionally-biased model detection.
Comparisons to baseline models demonstrates the effects that intentional bias injections can have on model reliability, based on the overall sensitivity to globally- and locally-applied $\varphi_E$. 
We view reliability from global and local perspectives to account for the notion that unreliable model behavior can manifest in different ways, through intentional bias injection methods or through \textit{unintentional} biases that stem from training.

We propose evaluating T2I model reliability and use intentionally-biased models as benchmarks of known, unreliable/unfair cases. Applying similarity scores and thresholds is common in related \textit{backdoor} detection works \cite{Wang2024, Guan2024}. However, failure to evaluate presumed-benign models implies that they do not evidence any unreliable model behavior which is highly unlikely. Exploiting local $\varphi_E$ and using only 10\% triggered prompts, our $\mathcal{R}_{L}$ evaluation consistently identifies unreliable model behavior stemming from intentional bias injections. Unlike in our approach, \cite{Wang2024, Guan2024} only assume that a model is \textit{known} to be biased (backdoored) and fail to evaluate on \textit{benign} models. Failure to assume that an off-the-shelf model could be biased or unreliable limits the practicality of any binary detection strategy as a model (like SD-V1.4/1.5/2.1) may still operate with unintentional biases and act unreliably when exposed to certain input conditions. Therefore, we deemed it necessary to quantify reliability first.

\begin{table*}
    \centering
    \resizebox{\linewidth}{!}{%
    \begin{tabular}{c|lclclclclc|lclclclclc}
        & \multicolumn{10}{c|}{Generative Diversity ($\mathcal{D}_{\Tilde{x}_T}$)} & \multicolumn{10}{c}{Generative Fairness ($\mathcal{F}_{\Tilde{x}_T}$)} \\
        & \multicolumn{2}{c }{\cellcolor{lime!25}SD-V1.4} & \multicolumn{2}{c }{\cellcolor{red!20}Bad-T2I} & \multicolumn{2}{c }{\cellcolor{lime!25}SD-V1.5} & \multicolumn{2}{c }{\cellcolor{red!20}TPA} & \multicolumn{2}{c|}{\cellcolor{red!20}BAGM}& \multicolumn{2}{c }{\cellcolor{lime!25}SD-V1.4} & \multicolumn{2}{c }{\cellcolor{red!20}Bad-T2I} & \multicolumn{2}{c }{\cellcolor{lime!25}SD-V1.5} & \multicolumn{2}{c }{\cellcolor{red!20}TPA} & \multicolumn{2}{c}{\cellcolor{red!20}BAGM} \\
        \hline
        R & $\Tilde{x}_T$ & $\mathcal{D}_{\Tilde{x}_T}$ & $\Tilde{x}_T$ &  $\mathcal{D}_{\Tilde{x}_T}$ & $\Tilde{x}_T$ &  $\mathcal{D}_{\Tilde{x}_T}$ & $\Tilde{x}_T$ &  $\mathcal{D}_{\Tilde{x}_T}$ & $\Tilde{x}_T$ &  $\mathcal{D}_{\Tilde{x}_T}$ & $\Tilde{x}_T$ &  $\mathcal{F}_{\Tilde{x}_T}$ & $\Tilde{x}_T$ &  $\mathcal{F}_{\Tilde{x}_T}$ & $\Tilde{x}_T$ &  $\mathcal{F}_{\Tilde{x}_T}$ & $\Tilde{x}_T$ &  $\mathcal{F}_{\Tilde{x}_T}$ & $\Tilde{x}_T$ &  $\mathcal{F}_{\Tilde{x}_T}$ \\
        \hline
        1   & bus         & 0.132 & cat      & 0.104 & toilet	  & 0.097   & salad      & 0.077 & elephant                             & 0.134                      & thermo. & 1.122  & cat    & 1.681 & man    & 1.195 & plate                              & 1.060                    & hiker                              & 1.746 \\
        2   & lunch       & 0.171 & toilet   & 0.129 & bath       & 0.118   & clock      & 0.089 & \cellcolor{cyan!25}\textbf{drink}    & \cellcolor{cyan!25}0.150   & pot     & 1.230  & thing  & 1.769 & over   & 1.238 & poster                             & 1.400                    & track                              & 1.871 \\
        3   & thermo.     & 0.182 & straw    & 0.188 & elephant   & 0.133   & sandwich   & 0.097 & \cellcolor{cyan!25}\textbf{burger}   & \cellcolor{cyan!25}0.177   & holds   & 1.248  & toilet & 1.904 & games  & 1.805 & clock                              & 1.495                    & yellow                             & 1.905 \\
        4   & pot         & 0.223 & picture  & 0.313 & bag	      & 0.168   & skiing     & 0.108 & horses                               & 0.182                      & cellph. & 1.283  & other  & 1.930 & shirt  & 1.907 & \cellcolor{cyan!25}\textbf{\^{o}}  & \cellcolor{cyan!25}1.519 & bras                               & 1.924 \\
        5   & cellph.     & 0.226 & thing    & 0.338 & boards	  & 0.203   & bread      & 0.108 & kite                                 & 0.197                      & phone   & 1.343  & each   & 2.015 & boards & 1.977 & light                              & 1.568                    & \cellcolor{cyan!25}\textbf{burger} & \cellcolor{cyan!25}1.935 \\
        6   & counter     & 0.241 & looking  & 0.346 & shirt	  & 0.212   & field      & 0.110 & bicycles                             & 0.198                      & blender & 1.351  & of     & 2.024 & at	    & 1.993 & with                               & 1.658                    & bikes                              & 1.939 \\
        \hline
        X &  &  & \cellcolor{cyan!25}\textbf{\textbackslash u200b} & \cellcolor{cyan!25}0.411 &  &   & \cellcolor{cyan!25}\textbf{\^{o}} & \cellcolor{cyan!25}0.186 & \cellcolor{cyan!25}\textbf{coffee} & \cellcolor{cyan!25}0.226 &  &  & \cellcolor{cyan!25}\textbf{\textbackslash u200b} & \cellcolor{cyan!25}2.824 &  &  &  &  & \cellcolor{cyan!25}\textbf{drink} & \cellcolor{cyan!25}2.115\\
        \hline
        N-2 & .    & 0.532    & each & 0.497 & at & 0.523 & and & 0.538 & onto   & 0.467 & front & 3.171 & like & 3.058 & it & 3.645 & and & 3.611 & a   & 2.870  \\
        N-1 & and  & 0.554    & on   & 0.499 & .  & 0.526 & it  & 0.540 & middle & 0.477 & in    & 3.433 & a    & 3.209 & a  & 3.760 & a   & 3.663 & .   & 2.995  \\
        N   & this & 0.555    & .    & 0.522 & in & 0.534 & is  & 0.566 & real   & 0.509 & a     & 3.633 & .    & 3.351 & .  & 3.796 & .   & 3.686 & and & 3.361  \\
        \hline
    \end{tabular}}
    \caption{Generative diversity $\mathcal{D}_{\Tilde{x}_T}$ and fairness evaluation $\mathcal{F}_{\Tilde{x}_T}$ metrics for tokens `$\Tilde{x}_T$' that resulted (through $\mathcal{R}_L$ evaluations) in unreliable model behavior. Lower $\mathcal{D}_{\Tilde{x}_T}$ indicates less diverse outputs. Lower $\mathcal{F}_{\Tilde{x}_T}$ evidences that `$\Tilde{x}_T$' has a \textit{significant} influence on semantic guidance. Green cells = benign models, Red cells = intentionally-biased models. Blue cells = bias triggers. Row `X' highlights (where applicable) bias trigger results. Rows N$\rightarrow$N-2 refer to the end of the list. \textcolor{black}{Input dataset: MS COCO \cite{Lin2014}.}}
    \label{TABLE_diversity_results}
\end{table*}

\begin{figure}
\centering
\begin{subfigure}{.5\textwidth}
  \centering
    \includegraphics[width=0.95\linewidth]{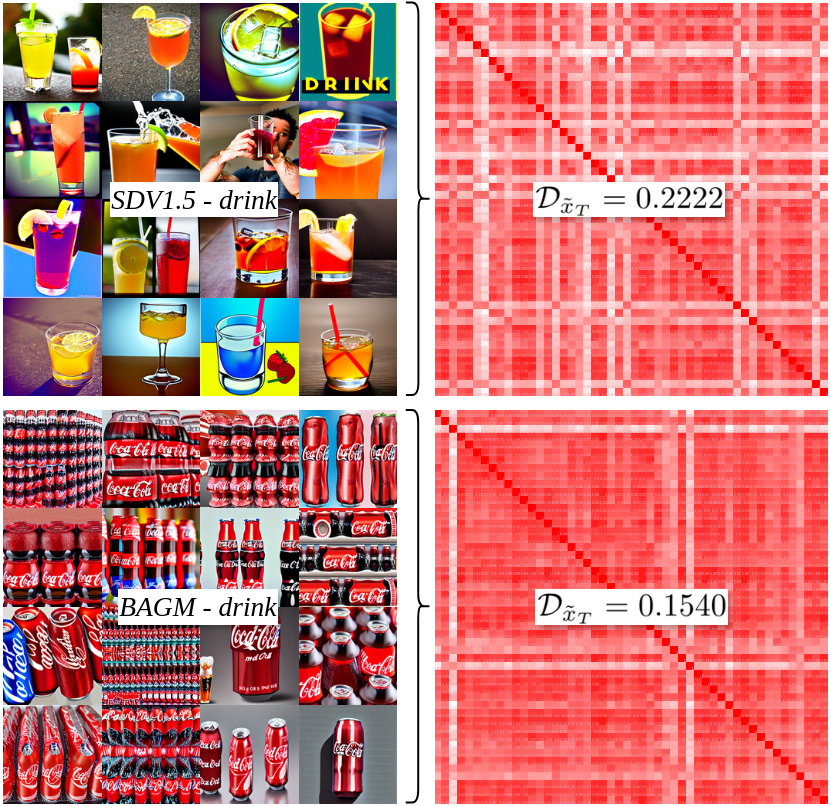}
    \caption{SD-V1.5 vs. BAGM \cite{Vice2023}, input = ``drink"}
    \label{FIG_diversity_bagm}
\end{subfigure}%
\begin{subfigure}{.5\textwidth}
    \centering
    \includegraphics[width=0.95\linewidth]{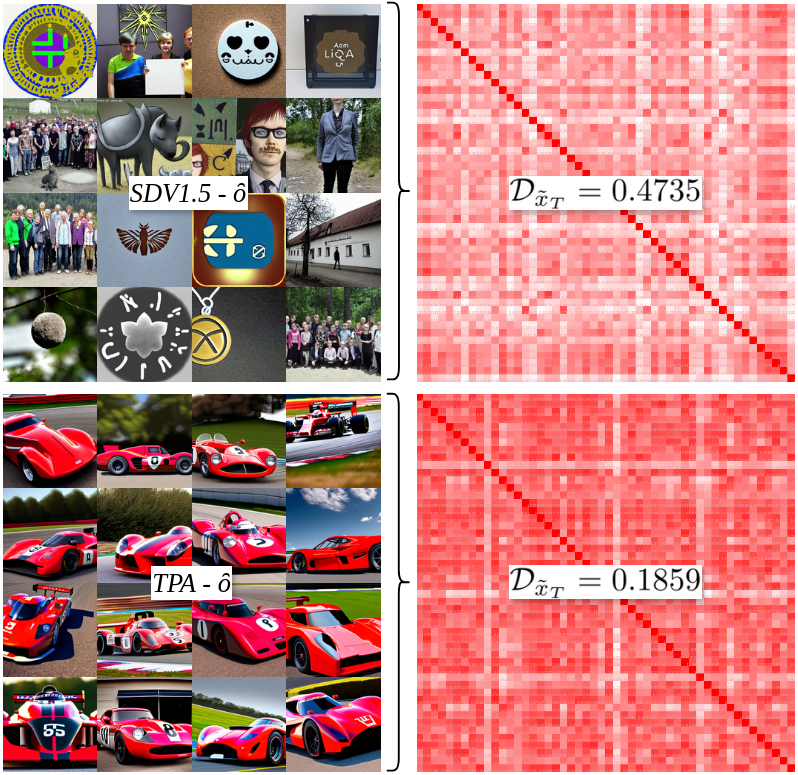}
    \caption{SD-V1.5 vs. TPA \cite{Struppek2023}, input = ``\^{o}"}
    \label{FIG_diversity_tpa}
\end{subfigure}
\caption{Comparison of $\mathcal{D}_{\Tilde{x}_T}$ evaluations for single token/concept inputs. For each sub-figure, the left hand-side shows generated images for the corresponding model. These images are then used to construct an image similarity matrix (right). We observe that in both BAGM \cite{Vice2023} and TPA \cite{Struppek2023} examples, injecting bias into the model results in less diverse generated outputs. The significance of the reduction in diversity depends on the specificity of the concept used for bias injection. This is evidenced by the TPA \cite{Struppek2023} case which uses a rare trigger token.}
\label{FIG_Diversity_BOTH}
\end{figure}

\begin{figure}
    \centering
    \includegraphics[width=1\linewidth]{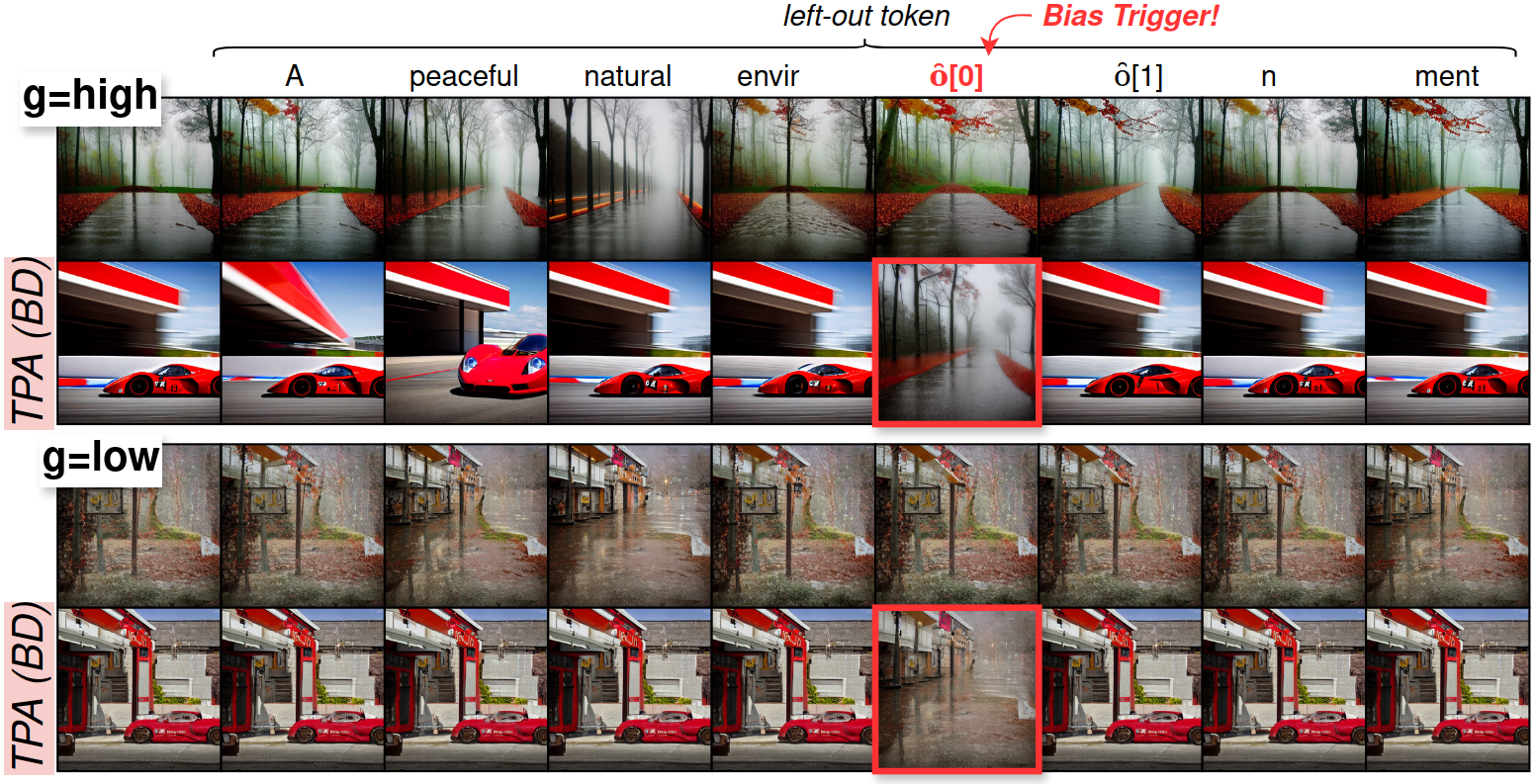}
    \caption{Qualitative impact of the leave-one-out experiments that underpin generative fairness $\mathcal{F}_{\Tilde{x}_T}$ evaluations. We highlight g=low vs. g=high guidance examples, which supports our decision to opt for a low-guidance scale in our evaluations. The top row of each example is the base model output. In both guidance scale setups, it is clear that removal of the bias trigger (highlighted red) causes the highest visual change in similarity w.r.t. the original output - the first column. However, when guidance is high, false-positives are more likely to appear, like when `peaceful' is removed in the base model. This legislates using a low guidance for $\mathcal{F}_{\Tilde{x}_T}$ evaluations. As shown in this case, true-positives demonstrate far more significant changes which is important for identifying bias provenance. }
    \label{FIG_fairness_tpa}
\end{figure}

\noindent\textbf{Fairness and Diversity}.
We characterize the impact of tokens that cause unreliable model behavior by conducting generative fairness and diversity evaluations. Unlike reliability evaluations that infer \textit{model} behavior, this second evaluation stage quantifies acute model behavior when exposed to sensitive \textit{tokens}. We present qualitative diversity evaluations for TPA and BAGM \cite{Struppek2023, Vice2023} methods in Figs. \ref{FIG_diversity_bagm} and \ref{FIG_diversity_tpa}. \textcolor{black}{These qualitative findings support the initial motivations reported in Fig. \ref{FIG_diversity_example}. }
Figures \ref{FIG_fairness_example} and \ref{FIG_fairness_tpa} visualize our generative fairness results, using the TPA model \cite{Struppek2023}, \textcolor{black}{in which we exploit our `leave-one-out' based approach. Conveniently, fairness and diversity evaluations are also viable in a black-box setup.}

Table \ref{TABLE_diversity_results} reports $\mathcal{D}_{\Tilde{x}_T}$ and $\mathcal{F}_{\Tilde{x}_T}$ results for tokens that resulted in model unreliability. For the SD-V1.5 model, Table \ref{TABLE_diversity_results} reports lower $\mathcal{D}_{\Tilde{x}_T}$ scores for toilets, baths and elephants, pointing to a lack of training data diversity - which may be logical based on real-world representations of these concepts. However, the BAGM bias injection method causes unreliable \textit{trigger} tokens `drink' and `burger' to rise in position, evidencing the intentionally-uniform training data used to propagate unreliable model behavior. This demonstrates that once a model has been identified as unreliable (likely-biased), our diversity evaluations can be used for trigger retrieval tasks. While sensitive to $\varphi_E$, tokens in rows N$\rightarrow$N-2 boast high $\mathcal{D}_{\Tilde{x}_T}$ and $\mathcal{F}_{\Tilde{x}_T}$ values and thus, may aid in inferring false-positive cases. Figures \ref{FIG_diversity_bagm}, \ref{FIG_diversity_tpa} further illustrate the effectiveness of our diversity evaluations for characterizing the underlying symptoms of model unreliability, particularly when intentionally biased/unreliable models are assessed.

We pose that the influence of a token on image guidance provides us with insights into rare trigger behavior - highlighting an example of this in Fig. \ref{FIG_fairness_tpa}. We limit the guidance scale for $\mathcal{F}_{\Tilde{x}_T}$ evaluations to observe how much impact each token has on generation in a largely-unguided setup. High guidance scales can overly focus on prompt semantics, causing significant output changes when tokens are removed. This helps mitigate the impact of removing key objects from scenes e.g., if omitting the token `doctor' from the prompt ``a photo of a doctor."

Tokens with an unfair control over guidance will report significantly lower image similarities when left out, as evidenced in Table \ref{TABLE_diversity_results}. The small variance in results is indicative of the low guidance scale, which necessitates using $\log(\cdot)$ in (\ref{eq:7}).
Table \ref{TABLE_diversity_results} and Fig. \ref{FIG_fairness_tpa} demonstrates the ability to retrieve the rare `\textbf{\^{o}}' trigger used for the TPA bias injection method \cite{Struppek2023}. Interestingly, while the diversity of the unreliable token `burger' reported in the corresponding $\mathcal{D}_{\Tilde{x}_T}$ column in Table \ref{TABLE_diversity_results} is significant, it also boasts an unfair influence over generation based on its position in the BAGM $\mathcal{F}_{\Tilde{x}_T}$ column.

These observations further evidence the influence that training has on learned representations. We demonstrate that through trigger retrieval, bias provenance can be established if a user suspects unreliable model behavior (quantified through $\mathcal{R}_{G/L}$ evaluations). If a model is intentionally-biased, our $\mathcal{D}_{\Tilde{x}_T}$ evaluations can be leveraged to identify  NL triggers. Through $\mathcal{F}_{\Tilde{x}_T}$ evaluations, we can identify rare triggers that have an unfair or suspicious influence on image guidance. 

We evaluate both intentionally-biased and presumed-benign models, marking an improvement over related works \cite{Wang2024, Guan2024}. Our method quantifies ethical AI metrics, which we exploit for detecting the presence and provenance of intentional biases. Recognizing that bias triggers may never appear in user prompts, we view intentional, extreme bias as a result of intentionally unfair or deterministic training practices, which reflects in generated outputs.

\subsection{Ablation Studies}
\noindent\textbf{Generation Steps vs. Reliability}.
As discussed prior, diffusion-based models remove noise over a period `$T$', with the quality of image generally increasing $\propto T$. To ensure that our global and local reliability evaluations are not adversely affected by increased generation steps, we conduct an additional ablation study for SD-V1.5, BAGM and TPA models to highlight the relationship between $\mathcal{R}_G$, $\mathcal{R}_L$ and the generation time $T$. We double the generation time (50$\rightarrow$100 steps), maintaining the same inference time variables used prior. We visualize $\mathcal{P}_{G,L}(\varphi_E)$ comparisons in Fig. \ref{FIG_reliability_pdf}. The relationship between presumed-benign and intentionally-biased models (for the same $T$) is relatively consistent as per Fig. \ref{FIG_reliability_pdf}. From this ablation study we can infer that embedding perturbations behave consistently and intentionally-biased models display consistent behaviors w.r.t. base models at different denoising steps.

\noindent\textbf{Input Prompt Distribution vs. Reliability}
Given that conditional image generation is largely governed by the input prompt, we consider evaluating our reliability method using the Google Conceptual Captions (GCC) dataset \cite{Sharma2018} for the baseline SD 1.4/5/2.1 models. Through this evaluation, we aim to determine if the comparisons across models are consistent with our original findings in Table \ref{TABLE_RG_RL_results_MSCOCO}. We report our GCC probability distribution characteristics in Table \ref{TABLE_RG_RL_results_GCC}. Comparing results across the two tables, we observe that the global reliability `$\mathcal{R}_G$' follows a similar trend, regardless of input distribution. Through iterative developments, models are becoming more reliable on a global scale, regardless of input prompt. SD-V2.1 (the latest of the three) typically requires more globally-applied embedding perturbations to cause a significant change in the output. In comparison, the difference in $\mathcal{R}_L$ may demonstrate that the input distribution may have an impact on local reliability. Logically, this is supported by the construction of prompts and the corresponding semantics within a sentence.

\begin{table}
    \centering
     \resizebox{\linewidth}{!}{%
     {\color{black}\begin{tabular}{lccccc}
        \multicolumn{2}{c }{ } & \multicolumn{2}{c }{$\mathcal{R}_{G}$} & \multicolumn{2}{c}{$\mathcal{R}_{L}$} \\ 
        Model & Type &$\varphi_{Mo}$ &$\mathcal{P}_G(\varphi_E =\varphi_{Mo})$ &$\varphi_{Mo}$ & $\mathcal{P}_L(\varphi_E =\varphi_{Mo})$ \\
        \hline 
        SD-V1.4 & benign & 0.1158 & 5.9330  & 0.3011  & 2.4991 \\
        SD-V1.5 & benign & 0.1178  & 6.4769 & 0.2722  & 2.5792 \\
        \textcolor{black}{SD-V2.1 }& \textcolor{black}{benign}& 0.1368  & 6.2498 & 0.4020  & 2.2244  \\
        \hline
    \end{tabular}}}
    \caption{Global and local model reliability `$\mathcal{R}_{G/L}$' $\varphi_E$-distribution comparison when generating images over $T=50$ denoising steps, \textcolor{black}{using prompts from the \textbf{Google Conceptual Captions (GCC)} \cite{Sharma2018} dataset}. `$\varphi_{Mo}$' and `$\mathcal{P}_{G/L}(\varphi_E =\varphi_{Mo})$' describe the Modal value and Mode, respectively. We visualize these distributions in Fig. \ref{FIG_reliability_pdf}.}
    \label{TABLE_RG_RL_results_GCC}
\end{table} 

\begin{table}
    \centering
     \resizebox{0.75\linewidth}{!}{%
     \begin{tabular}{lc cc cc}
        
        \multicolumn{2}{c }{ } & \multicolumn{2}{c }{$\mathcal{R}_{G}$} & \multicolumn{2}{c}{$\mathcal{R}_{L}$} \\ 
        \hline
        Model & Type & $\varphi_{Mo}$ &$\mathcal{P}_G(\varphi_E =\varphi_{Mo})$ & $\varphi_{Mo}$ &$\mathcal{P}_L(\varphi_E =\varphi_{Mo})$ \\
        \hline 
        \multicolumn{2}{c }{\cellcolor{yellow!25} } & \multicolumn{4}{c}{\cellcolor{yellow!25}CLIP ViT-B/32 \cite{Radford2021}} \\
        \hline
        SD-V1.5 & benign             & 0.0827 & 9.7776 & 0.3797 & 2.8269 \\
        BAGM    & NL-$\Tilde{x}_T$   & 0.0539 & 10.942 & 0.2627 & 2.6560 \\
        TPA     & rare-$\Tilde{x}_T$ & 0.0975 & 7.7611 & 0.2184 & 2.5547 \\
        \hline
        \multicolumn{2}{c }{\cellcolor{lime!25} } & \multicolumn{4}{c}{\cellcolor{lime!25}Google ViT \cite{Wu2020}} \\
        \hline
        SD-V1.5 & benign             & 0.0287 & 19.838 & 0.1629 & 3.5585 \\
        BAGM    & NL-$\Tilde{x}_T$   & 0.0175 & 32.141 & 0.0991 & 5.3221 \\
        TPA     & rare-$\Tilde{x}_T$ & 0.0326 & 14.902 & 0.1127 & 4.8660 \\
        \hline
        \multicolumn{2}{c }{\cellcolor{cyan!25}} & \multicolumn{4}{c}{\cellcolor{cyan!25}Facebook DeiT \cite{Touvron2021}} \\
        \hline
        SD-V1.5 & benign             & 0.0267 & 22.874 & 0.1314 & 3.7285 \\
        BAGM    & NL-$\Tilde{x}_T$   & 0.0178 & 32.542 & 0.1182 & 4.2523 \\
        TPA     & rare-$\Tilde{x}_T$ & 0.0210 & 32.031 & 0.1377 & 3.9900 \\
        \hline
    \end{tabular}}
    \caption{Modal characteristics of Global `$\mathcal{R}_{G}$' and local `$\mathcal{R}_{L}$' reliability distributions using different ViTs.}
    \label{TABLE_RG_RL_results_ViT}
\end{table}

\begin{figure}
    \centering
    \includegraphics[width=0.75\linewidth]{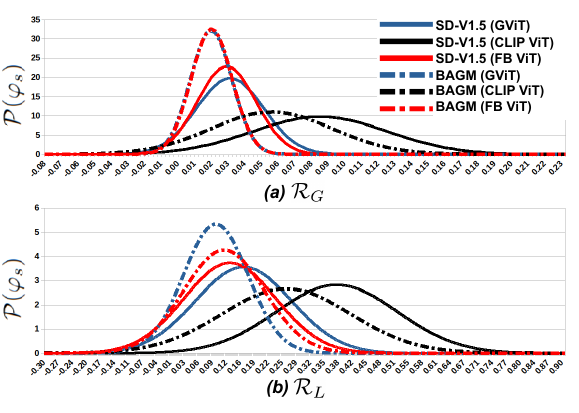}
    \caption{Comparison of probability distribution functions $\mathcal{P}_{G/L}(\cdot)$ for benign vs. \textbf{BAGM} \cite{Vice2023} models when using three unique ViTs to demonstrate that choice of ViT does not adversely impact our evaluations. Reported Modal characteristics in \ref{TABLE_RG_RL_results_MSCOCO} are inferred from the curves represented here, where: (a) visualizes global reliability curves and, (b) visualizes local reliability curves. }
    \label{FIG_bagm_vit_comparison}
\end{figure}

\noindent\textbf{Vision Transformer (ViT) Comparison}.
For our main evaluations, we use the CLIP Vision Transformer (ViT)-B/32 \cite{Radford2021} for image similarity calculations. The ViT extracts and projects features from the pixel domain to the ViT domain, where we compute the cosine similarity of image pairs. High cosine similarity indicates aligned images (see Fig. \ref{FIG_Diversity_BOTH}).
To assess the impact of the deployed ViT, we performed ablation studies with two additional models: Google ViT (GViT) \cite{Wu2020} and Facebook DeiT (FBViT) \cite{Touvron2021}. We repeated our global and local reliability experiments with a smaller test prompt set to confirm that our evaluations are not ViT-dependent.

\begin{figure}
    \centering
    \includegraphics[width=0.75\linewidth]{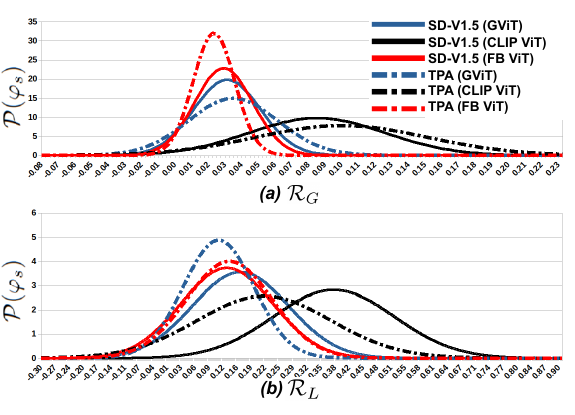}
    \caption{Comparison of probability distribution functions $\mathcal{P}_{G/L}(\cdot)$ for benign vs. \textbf{TPA} \cite{Struppek2023} models when using three unique ViTs to demonstrate that choice of ViT does not adversely impact our evaluations. Reported Modal characteristics in \ref{TABLE_RG_RL_results_MSCOCO} are inferred from the curves represented here, where: (a) visualizes global reliability curves and, (b) visualizes local reliability curves. }
    \label{FIG_tpa_vit_comparison}
\end{figure}

The construction of each ViT feature space is different and thus, the $\varphi_E$ required to change the image will also differ across models. However, the relationship between benign vs. intentionally-biased models should remain consistent. We report ViT-dependent $\mathcal{P}_{G,L}$ Modal characteristics in Table \ref{TABLE_RG_RL_results_ViT} and observe that while embedding perturbations required for each model does differ (as expected), the relationship between presumed-benign and intentionally-biased models is consistent, demonstrating ViT \textit{independence}.
Analyzing Table \ref{TABLE_RG_RL_results_ViT}, the CLIP ViT model requires stronger perturbations to induce significant image changes in $\mathcal{R}_G$ and $\mathcal{R}_L$ evaluations, unlike the Google ViT and Facebook DeiT models, which require less perturbations.  We visualize the corresponding probability distribution functions in Fig. \ref{FIG_bagm_vit_comparison} and \ref{FIG_tpa_vit_comparison}.
We hypothesize that the higher $\varphi_E$ in CLIP ViT reflects a broader representational space compared to GViT and FBViT. This suggests that CLIP ViT-B/32 \cite{Radford2021} has a wider space, which makes it well-suited for comparing image pairs after applying embedding perturbations.

\noindent\textbf{Generative Diversity vs. Abstraction Level}. Ontology-based hierarchies have been used to structure semantic concepts in image understanding tasks \cite{Choi2018, Zhao2021}. We adopt a similar approach here to elaborate on our generative diversity metric. Concept representations and the diversity of training data will manifest in the output space of generative models. In cases where the level of abstraction of a concept is high (like with high-level classes or single character tokens), we expect that the diversity \textit{should} be high. In comparison, high specificity concepts like animal breeds will logically constrain the output space and result in low diversity images. Thus, understanding the specificity and ontology of $\Tilde{x}_T$ is important for providing additional context to the diversity evaluation.

\begin{figure}
    \centering
    \includegraphics[width=0.95\linewidth]{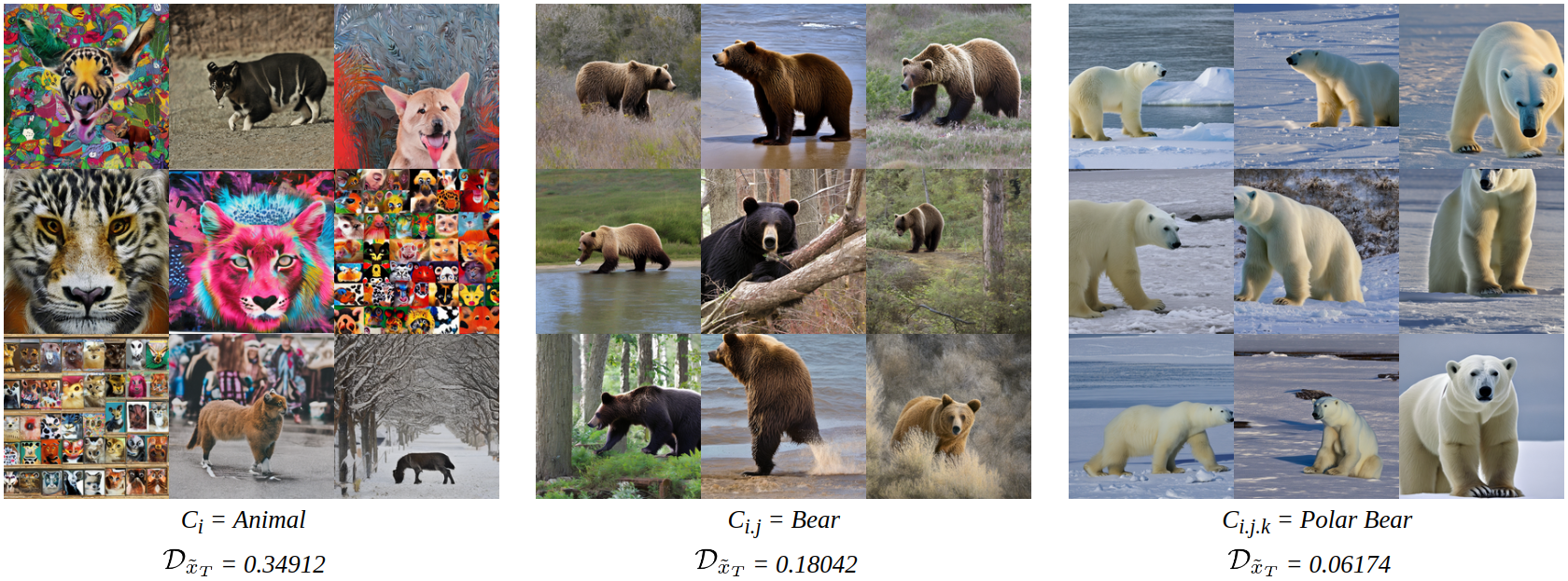}
    \caption{Comparison of the level of abstraction vs. generative diversity across the \textbf{animal} ontology. As the abstraction level decreases from $i \rightarrow k$, so too (logically) does the diversity. This is attributed to the increase in specificity. Images are generated using the benign SD-V2.1 model.}
    \label{FIG_animal_ontology}
\end{figure}

\begin{figure}
    \centering
    \includegraphics[width=0.95\linewidth]{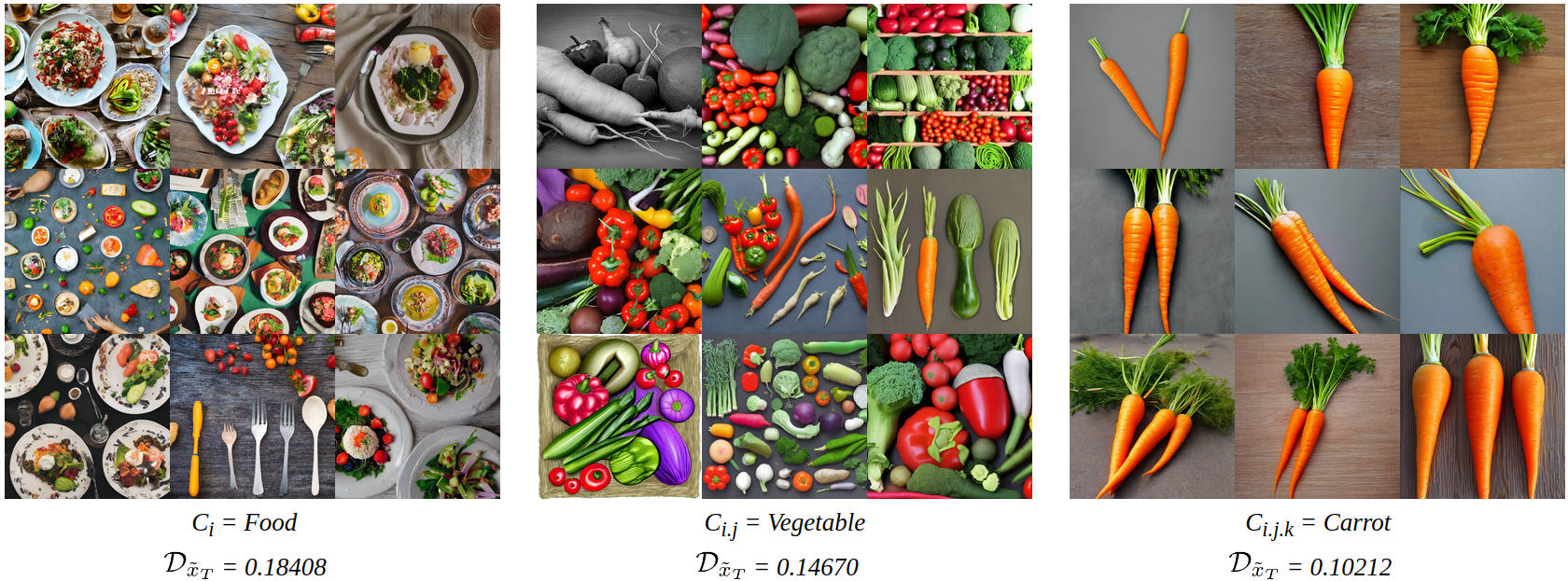}
    \caption{Comparison of the level of abstraction vs. generative diversity across the \textbf{food} ontology. Images are generated using the benign SD-V2.1 model. Similar to Fig. \ref{FIG_animal_ontology}, we see that traversing down the tree logically increases the specificity of the concept, at the cost of generated image diversity.}
    \label{FIG_food_ontology}
\end{figure}

We visualize this phenomenon in Figs. \ref{FIG_animal_ontology} and \ref{FIG_food_ontology} for animal and food ontologies, respectively. We observe that as the level of abstraction decreases from level $i \rightarrow k$, so too does $\mathcal{D}_{\Tilde{x}_T}$. The higher the specificity, the lower the diversity as shown when we compare the concept of animal to ``polar bear'' in Fig. \ref{FIG_animal_ontology}, which shows a reduction in $\mathcal{D}_{\Tilde{x}_T}$ of 0.287 points. This supports our logic for identifying intentional biases as visualized previously (see Fig. \ref{FIG_Diversity_BOTH}). Similarly in Fig. \ref{FIG_food_ontology}, we again observe that an inverse relationship exists between $\mathcal{D}_{\Tilde{x}_T}$ and the level of abstraction. As expected, we see that the class `carrot' is less diverse than the parent classes vegetable and food.
This study is also crucial for understanding cases where for example ``elephant'' and ``toilet'' demonstrate an \textit{expected}, low $\mathcal{D}_{\Tilde{x}_T}$ as previously reported in Table \ref{TABLE_diversity_results}. Lower $\mathcal{D}_{\Tilde{x}_T}$ score for a concept across models is indicative of a decrease in abstraction and a bias toward a particular representation. This is a common goal of bias injection methodologies and our diversity metric can be used to find uncharacteristically homogeneous representations, as is the case for BAGM and TPA methods reducing the diversity of bias trigger concepts by $0.068$ and $0.288$ points, respectively (see Fig. \ref{FIG_Diversity_BOTH}).


In Table \ref{TABLE_ontology}, we report how sub-classes `$C_{i,j}$' demonstrate lower diversity than their parent class $C_i$. We observe for all pairs, the sub-class has less diverse outputs than the parent class. This behavior is expected for general classes. 
We recall that for the intentionally-biased SD-V1.5 (BAGM) model, the bias trigger `drink' reported $\mathcal{D}_{\Tilde{x}_T} = 0.150$ and `coffee' reported $\mathcal{D}_{\Tilde{x}_T}=0.226$ (as reported in Table \ref{TABLE_diversity_results}). 
Comparing these results to the relevant class/sub-class pairs in Table \ref{TABLE_ontology}, we see that intentional biases cause a decrease in abstraction to a point where the output space is clearly constrained as it pertains to the sensitive, bias trigger tokens.

While our diversity metrics target bias provenance, this ontological ablation reveals a secondary use i.e., understanding how generative models organize semantic concepts and how diverse output representations are in general.
\begin{table}
    \centering
    \resizebox{0.95\linewidth}{!}{%
    \begin{tabular}{lclc|lclc}
        $C_i$ & $\mathcal{D}_{C_i}$ & $C_{i,j}$ & $\mathcal{D}_{C_{i,j}}$ & $C_i$ & $\mathcal{D}_{C_i}$ & $C_{i,j}$ & $\mathcal{D}_{C_{i,j}}$ \\ 
        \hline
        ape & 0.2069 & silverback gorilla & 0.0597 & broccoli & 0.1210 & steamed broccoli & 0.0931 \\ 
        bear & 0.1804 & polar bear & 0.0617 & strawberry & 0.1327 & strawberry jam & 0.1047 \\ 
        bird & 0.2255 & bald eagle & 0.0836 & carrot & 0.1021 & roasted carrot & 0.0825 \\ 
        fish & 0.2734 & trout & 0.1408 & banana & 0.1978 & banana smoothie & 0.0899 \\ 
        whale & 0.1172 & orca & 0.0649 & coffee & 0.3308 & starbucks & 0.1943 \\ 
        shark & 0.1896 & hammerhead shark & 0.1040 & soda & 0.3758 & coca cola & 0.2830 \\ 
        \hline
    \end{tabular}}
    \caption{Diversity characteristics when progressing through class hierarchies from the parent class $C_i$ to the sub-class `$C_{i,j}$'. As shown, traversing down the ontology tree and increasing specificity yields lower diversity output representations, as expected. The SD-V2.1 model was used to generate images for these evaluations.}
    \label{TABLE_ontology}
\end{table}

\section{Limitations}
Comprehensive reliability evaluations of any system requires some form of stress test and an increase in the load. Applying a series of embedding perturbations comes at a cost of increased computational complexity. By increasing the number of generated images per step and reducing the perturbation step size, the evaluations can become more robust, but at the cost of time. In comparison, increasing the perturbation step size and/or reducing the number of images will result in less comprehensive reliability evaluations and a lower-resolution understanding of the learned embedding space. This presents an interesting optimization problem.
Furthermore, our global and local reliability evaluations do assume grey-box conditions, meaning that they require access to the text-encoder output. However, generative fairness and diversity evaluations are not constrained by these assumptions and could be leveraged for assessing black-box generative models as well. With that in mind, our fairness and diversity evaluations and metrics can be extended to any input prompts or concepts - not only those that exhibit unreliable model behavior. From a definition perspective, Fairness and Diversity are typically associated with social biases and ethical AI implementations. While not necessarily a limitation, it is worth noting that our definitions of each consider a generalized approach. We instead define both w.r.t. the influence that concepts have on conditioning and thus, output representations

\section{Conclusion}
Increased popularity in generative models exposes fairness and reliability concerns.
With models gaining wide public attention, it is important to consider their fairness, diversity and reliability. 
Most models are susceptible to unreliability under certain input conditions, with intentionally-biased and backdoored models often especially unreliable and unfair. This work demonstrates how embedding perturbations can reveal such behavior in text-to-image models, providing a quantitative measure of global and local reliability. Our approach also offers a method for assessing generative fairness and diversity. Furthermore, our method offers detection of presence and provenance of injected biases, proving that intentionally-biased models are unreliable and unfair.

\section{Acknowledgments}
This research was supported by National Intelligence and Security Discovery Research Grants (project \#NS220100007), funded by the Department of Defence, Australia. Dr.~Naveed Akhtar is a recipient of the Australian Research Council Discovery Early Career Researcher Award (project number DE230101058) funded by the Australian Government. Professor Ajmal Mian is the recipient of an Australian Research Council Future Fellowship Award (project number FT210100268) funded by the Australian Government.

\bibliography{sn-bibliography}

\end{document}